\documentclass[a4paper,11pt]{article}
\usepackage[utf8]{inputenc}
\usepackage[T1]{fontenc}

\usepackage{libertine} 
\usepackage{inconsolata} 
\usepackage{needspace}

\RequirePackage{etex} 

\usepackage{lineno}
\usepackage{url}
\usepackage{hyperref}  
\usepackage[numbers,square]{natbib}
\usepackage{amsmath}
\usepackage{amssymb}
\usepackage{booktabs}
\usepackage{amsfonts,amsthm}
\usepackage{thmtools,thm-restate}
\usepackage{tabularx,lipsum, environ}
\usepackage{bbm,hyperref}
\hypersetup{
	colorlinks=true,
    linkcolor={red!50!black},
    citecolor={blue!50!black},
    urlcolor={blue!80!black},
	bookmarksopen=true,
	bookmarksnumbered,
	bookmarksopenlevel=2,
	bookmarksdepth=3
}
\usepackage[capitalize]{cleveref}
\usepackage{graphicx,float}
\usepackage{linegoal}
\usepackage{caption}
\usepackage{subcaption}
\usepackage{complexity}
\usepackage{enumerate}
\usepackage{multirow}
\usepackage{cleveref}
\usepackage{enumitem,linegoal}
\usepackage{todonotes}
\usepackage{amsthm,color}
\usepackage{amsmath}
\usepackage{graphics}
\usepackage{fullpage}
\usepackage{amssymb}
\usepackage{graphicx}
\usepackage{lipsum}
\usepackage{comment}
\usepackage{caption,enumerate}

\theoremstyle{remark}

\theoremstyle{definition}

\theoremstyle{plain}
\newtheorem{open problem}{Open problem}

\let\plainqed\qedsymbol

\newcommand\restr[2]{{
		\left.\kern-\nulldelimiterspace 
		#1 
		\littletaller 
		\right|_{#2} 
}}

\newcommand{\Romannum}[1]{\uppercase\expandafter{\romannumeral #1\relax}}

\newcommand{\littletaller}{\mathchoice{\vphantom{\big|}}{}{}{}}

\usepackage{tikz}
\usetikzlibrary{arrows}
\usetikzlibrary{arrows.meta}
\usetikzlibrary{decorations.pathmorphing}
\usetikzlibrary{decorations.markings}
\usetikzlibrary{calc}
\usetikzlibrary{positioning}
\tikzset{
	circ/.style = {circle,draw,fill,inner sep=1.3pt},
	mcirc/.style = {circle,draw,fill,inner sep=1pt},
	circR/.style = {circle,draw=red,fill=red,text=red,inner sep=1.3pt},
	circb/.style = {circle,draw=blue,fill=blue,text=blue,inner sep=1.1pt},
	circr/.style = {circle,draw=red,fill=red,inner sep=1pt},
	scirc/.style = {circle,draw,fill,inner sep=.8pt},
	invisible/.style = {draw=none,inner sep=0pt,font=\tiny},
	nonedge/.style={decorate,decoration={snake,amplitude=.3mm,segment length=1mm},draw}
}

\usepackage{authblk}



\title{\LARGE Explainable Graph-theoretical Machine Learning\\
\vspace{.5mm} 
with Application to Alzheimer's Disease Prediction}

\author[1,2$\dag$]{Narmina Baghirova} 
\author[2,3]{Duy-Thanh V\~u}
\author[2,3]{Duy-Cat Can}
\author[2,3]{Christelle Schneuwly Diaz}

\author[2,3]{\authorcr Julien Bodlet}
\author[2]{Guillaume Blanc}
\author[2,3]{Georgi Hrusanov}
\author[1]{Bernard Ries}
\author[2,3$\ddag$]{and Oliver Y. Ch\'en}
\author[ ]{\qquad \qquad for the Alzheimer's Disease Neuroimaging Initiative\thanks{Data used in the preparation of this article are from the Alzheimer's Disease Neuroimaging Initiative (ADNI) database (\url{http://adni.loni.usc.edu}). As such, the investigators within the ADNI contributed to the design and implementation of ADNI and/or provided data but did not participate in the analysis or writing of this paper. A complete list of ADNI investigators is available at: \url{http://adni.loni.usc.edu/wp-content/uploads/how_to_apply/ADNI_Acknowledgement_List.pdf}.}}

  \affil[1]{Department of Informatics, University of Fribourg, Fribourg, Switzerland}
  \affil[2]{Platform of Bioinformatics, Lausanne University Hospital (CHUV), Lausanne, Switzerland}
  \affil[3]{Faculty of Biology and Medicine, University of Lausanne, Lausanne, Switzerland
  \vspace{2mm}
  \newline
\textsuperscript{\dag}\href{mailto:narmina.baghirova@unifr.ch}{narmina.baghirova@unifr.ch}; \textsuperscript{\ddag}\href{mailto:olivery.chen@chuv.ch}{olivery.chen@chuv.ch}.}

\date{}

\begin{document}
\maketitle

\vspace{-4mm}
\begin{sloppypar}
\begin{abstract}
Dementia affects over 55 million people worldwide, projected to reach 139 million by 2050, with Alzheimer's disease (AD) accounting for 60--70\% of cases. AD is associated with disruptions in metabolic brain connectivity. Detecting these disruptions early is crucial for AD management. Fluorodeoxyglucose positron emission tomography (FDG-PET) is a useful tool for identifying such impairments. However, most studies rely on group-level analyses or thresholding, potentially masking individual differences and overlooking weaker yet biologically critical brain connections. Moreover, AD prediction largely focuses on univariate rather than multivariate outcomes. To address this, we introduce \textit{explainable graph-theoretical machine learning} (XGML), a framework for constructing individual metabolic brain graphs and identifying subgraphs most predictive of multivariate disease-related outcomes. Using Alzheimer's Disease Neuroimaging Initiative (ADNI) FDG-PET data, we compared six graph representations against three non-graph baselines, each with six machine learning models using repeated stratified 3-fold cross-validation (10 repeats). The best configuration combined kernel density estimation with Hellinger distance and random forest. Across eight cognitive scores, it reached an overall Fisher-z-averaged Pearson correlation of $r=0.595$, with the strongest predictive performance for ADAS13 ($r=0.67$), ADAS11 ($r=0.65$), and ADASQ4 ($r=0.62$). We identified key edges that were jointly but differentially predictive across outcomes, suggesting their potential as network biomarkers of cognitive decline. Preliminary external feasibility validation on an OASIS3 cohort yielded weak predictive     performance for CDRSB ($r=0.26$) and MMSE ($r=0.18$), likely reflecting cohort, protocol, and diagnostic differences. Overall, our results suggest the promise of graph-theoretical machine learning for biomarker discovery, disease prediction, and understanding the neural mechanisms underlying AD.

\end{abstract}
\end{sloppypar}

\newpage
\section{Introduction}
\label{sec:introduction}
Alzheimer's disease (AD) is the predominant form of dementia \cite{who_dementia_2023}. It affects millions of individuals worldwide and leads to significant and growing global challenges \cite{alzheimers_research_uk_2021}. As the disease progresses, it impairs memory, cognition, and daily functioning, reducing patients’ quality of life and ultimately contributing to death. It places a substantial burden on caregivers and healthcare systems \cite{snm_alzheimers}. The limitations of current therapies \cite{passeri2022alzheimers}, combined with the observation that treatments are most effective in the early or middle stages of AD \cite{nia_alzheimers_treatment}, highlight the critical need for early detection and accurate assessment of disease severity and its progression.

Central to evaluating disease severity and cognitive impairment in AD are cognitive assessments \cite{McKhann2011}. Cognitive scores provide quantifiable measures of cognitive impairment and are widely used to characterize clinical manifestations of the disease. Biomarkers, in turn, provide complementary information about the biological processes underlying AD \cite{Dubois2023}. Establishing the relationship between such biological alterations and cognitive impairment is therefore important for understanding how disease-related changes in the brain are reflected in cognitive function. For example, cerebrospinal fluid tau and amyloid-beta provide valuable molecular and pathological information about AD \cite{taubetabiom}, while neuroimaging modalities offer complementary insights into structural and functional brain changes associated with disease progression. In particular, fluorodeoxyglucose positron emission tomography (FDG-PET) offers insights into functional changes in the brain as AD progresses. Specifically, FDG-PET measures cerebral glucose metabolism, which is related to neuronal activity \cite{FDG_PET}, enabling detection of functional impairments that may appear before magnetic resonance imaging (MRI)-detectable atrophy \cite{Mosconi2006}. Moreover, FDG-PET mirrors cognitive performance \cite{Henkel2020}, making it particularly valuable for AD assessment. Indeed, prior work highlights FDG-PET as a powerful tool for cognitive assessment \cite{PET_pred,Shokouhi2013}. 

While machine learning frameworks have been widely used to perform AD-related predictions, the existing methods tend to focus on univariate outcomes, such as disease status \cite{univariate1,univariate2} and individual cognitive measures \cite{univariate4,univariate6}. 
More recently, variational graph autoencoder frameworks have been proposed to learn latent representations from PET-derived brain graphs by integrating node features and weighted edges. For example, graph isomorphism networks with edge features (GINE)-based variational graph autoencoders have demonstrated strong performance in AD classification tasks on Alzheimer's Disease Neuroimaging Initiative (ADNI) data \cite{11222021}. From another perspective, network neuroscience, which applies graph theory to investigate brain networks, has made important strides thanks to the increasing availability of neuroimaging data \cite{bassett2017network}. Recently, it has helped reveal structural changes in the brain \cite{John2017}, identify potential network biomarkers \cite{gomez_ramirez2014}, and advance disease prediction methods \cite{Wei2016} in AD research. A common way to build brain graphs using neuroimaging data is to use vertices to represent brain regions of interest (ROIs) and edges to capture the relationships between ROIs. The resulting graphs may be unweighted, often following thresholding, or weighted, with statistical measures used to assign edge weights representing relationships between ROIs. As discussed in Section~\ref{sec:discussions}, graph-based studies on FDG-PET data commonly involve using group-level analyses \cite{FDG-PET_most_studies} and thresholding techniques; examples of such studies include \cite{PET_classification, SanabriaDiaz2013}. The use of group-level analyses can mask individual differences \cite{indivdiff}, while thresholding may discard weaker yet biologically relevant edges, as disruptions between brain regions in AD are not yet fully understood \cite{critedges}.

To address these limitations, we introduce \textit{explainable graph-theoretical machine learning} (XGML). In brief, XGML constructs individual metabolic brain graphs, uses their edge weights as features to predict multivariate disease-related outcomes, and identifies subgraphs most predictive of these outcomes. To evaluate the framework, we use FDG-PET scans from ADNI to construct individual metabolic brain graphs for each subject. We compare several graph representations and machine learning models within a unified cross-validation framework and benchmark them against non-graph baselines. Among all evaluated representation–model combinations, KDE+Hellinger graph construction combined with random forest achieved the highest predictive performance for the multivariate prediction of eight AD-related cognitive scores. Finally, using permutation feature importance, XGML identifies key edges that are jointly but differentially predictive of eight AD-related outcomes. We also perform a preliminary external feasibility validation on an independent OASIS3 cohort using the shared cognitive measures. 

While the present study evaluates XGML using FDG-PET data, the framework is not restricted to this imaging modality and could, in principle, be applied to other neuroimaging modalities, such as structural MRI and amyloid-PET.

To the best of our knowledge, no existing machine learning frameworks have used FDG-PET data to jointly predict the set of AD-related cognitive scores considered in this study. Prior work has primarily focused on univariate prediction using alternative data modalities. For example, speech-based approaches have explained up to 47\% of the variance in MMSE scores ( $R^2$ = 0.468 )~\cite{univariate4}, while resting-state functional MRI (rs-fMRI)-based models achieved a Spearman’s correlation of \( 0.49 \) for ADAS11 prediction~\cite{univariate6}. Similarly, deep learning models using structural MRI have reported correlations of \( r = 0.669 \) for CDRSB and \( r = 0.605 \) for MMSE~\cite{LI2025110579}, and \( r = 0.68 \) for ADAS13~\cite{bhagwat2019ann_alzheimers}. Furthermore, in other related studies where graphs were constructed from FDG-PET scans using techniques such as thresholding or group-level correlation, the analysis typically focused on applying standard graph-theoretical metrics, such as network efficiency, clustering coefficient, and small-worldness, to distinguish between different patient groups~\cite{tuan2025fdg}. In contrast, in this paper, we use features extracted from the constructed graphs as input to a predictive model.

The main contributions of this paper are summarized as follows: 
(1) We introduce a novel, explainable framework for constructing subject-specific metabolic brain graphs from FDG-PET data and using their connectivity patterns for multivariate prediction of AD-related cognitive scores.
(2) We evaluate multiple graph representations against non-graph baselines across different machine learning models, demonstrating that predictive performance depends substantially on how subject-specific brain graphs are constructed.
(3) XGML predicts eight AD-related cognitive scores, achieving correlations between observed and predicted values ranging from approximately $r=0.47$ to $r=0.67$ across outcomes.
(4) XGML identifies outcome-specific predictive edges as well as edges shared across multiple cognitive scores. The identified patterns show correspondence with brain regions implicated in the respective cognitive domains, supporting their neurobiological plausibility and motivating further investigation as candidate network biomarkers.
(5) To promote transparency and reproducibility, we make the code for graph construction, prediction, preliminary external feasibility validation, and feature importance analysis publicly available at: \textcolor{blue}{\href{https://github.com/THELabTop/XGML}{github.com/THELabTop/XGML}}.

\section{Materials and methods}
\label{sec:materialsandmethods}
In this article, we used data from ADNI (\href{https://adni.loni.usc.edu/}{adni.loni.usc.edu}) and the Open Access Series of Imaging Studies (OASIS3)~\cite{oasis3}.

The ADNI was launched in 2003 as a public-private partnership, led by Principal Investigator Michael W. Weiner, MD. The original goal of ADNI was to test whether serial MRI, positron emission tomography (PET), other biological markers, and clinical and neuropsychological assessment can be combined to measure the progression of mild cognitive impairment (MCI) and early AD. The current goals include validating biomarkers for clinical trials, improving the generalizability of ADNI data by increasing diversity in the participant cohort, and providing data concerning the diagnosis and progression of Alzheimer’s disease to the scientific community. For up-to-date information, see \href{https://adni.loni.usc.edu/}{https://adni.loni.usc.edu/}.

The OASIS3 cohort is a longitudinal neuroimaging, clinical, and cognitive dataset designed to support research on normal aging and AD, providing multimodal imaging (including MRI and PET) along with associated clinical and cognitive assessments.

All data were accessed and used in accordance with their respective Data Use Agreements.

\subsection{ADNI and OASIS3 Subjects}


For the main analysis, we constructed a balanced ADNI cohort consisting of 600 subjects, including 200 cognitively normal (CN), 200 MCI, and 200 AD subjects, selected based on the availability of FDG-PET scans and cognitive assessments for all eight target outcomes. To ensure consistency between imaging and cognition, each scan was matched to the closest cognitive assessment, preferring an interval of at most 90 days and relaxing this to 120 days only when necessary. The resulting scan–assessment interval ranged from 0 to 91 days (mean 2.65). Demographics are summarized in Table~\ref{tab:adni_demographics}.

\begin{table}[ht!]
\centering
\caption{Demographic characteristics of the ADNI cohort. No. = number of subjects. Age = mean age ($\pm$ standard deviation).}
\label{tab:adni_demographics}
\vspace{0.3em}

\begin{tabular}{lcccccc}
\toprule
& \multicolumn{2}{c}{\textbf{CN}}
& \multicolumn{2}{c}{\textbf{MCI}}
& \multicolumn{2}{c}{\textbf{AD}} \\
\cmidrule(r){2-3}
\cmidrule(r){4-5}
\cmidrule(r){6-7}
& \textbf{No.} & \textbf{Age}
& \textbf{No.} & \textbf{Age}
& \textbf{No.} & \textbf{Age} \\
\midrule
\textbf{Male} & 105 & 74.69 $\pm$ 5.51 & 128 & 74.43 $\pm$ 7.54 & 116 & 75.91 $\pm$ 7.64 \\
\textbf{Female} & 95 & 74.07 $\pm$ 5.47 & 72 & 72.08 $\pm$ 7.51 & 84 & 73.13 $\pm$ 7.62 \\
\midrule
\textbf{Total} & 200 & 74.39 $\pm$ 5.49 & 200 & 73.58 $\pm$ 7.59 & 200 & 74.75 $\pm$ 7.74 \\
\bottomrule
\end{tabular}

\end{table}

For preliminary external feasibility validation, we used an independent OASIS3 cohort~\cite{oasis3}, selecting subjects with an FDG-PET scan and the two shared measures (CDRSB and MMSE), yielding 101 subjects. Given the limited sample, no interval constraint was imposed. As exact assessment dates are unavailable, the scan–assessment interval was approximated from the difference between the subject's age at scan and at assessment, giving a coarse estimate ranging from 0 to 591.70 days (mean 158.29). This age-difference approach provides only a coarse approximation of the true interval between FDG-PET acquisition and cognitive assessment. Also, it is important to note that diagnostic labels in OASIS3 are not directly comparable to those in ADNI. Diagnostic labels were derived from the \texttt{dx\_chosen\_label} field in the OASIS3 Uniform Data Set (UDS), mapping \textit{Cognitively normal} to CN, MCI-related diagnoses to MCI, and AD-related diagnoses to AD, yielding an imbalanced cohort comprising 94 CN, 5 MCI, and 1 AD subject. Moreover, one subject had no available diagnostic label and was included in the analysis but excluded from the group-wise demographic summary reported in Table~\ref{tab:oasis_demographics}. 

\begin{table}[ht!]
\centering
\caption{Demographic characteristics of the OASIS3 cohort. No. = number of subjects. Age = mean age ($\pm$ standard deviation).}
\label{tab:oasis_demographics}
\vspace{0.3em}

\begin{tabular}{lcccccc}
\toprule
& \multicolumn{2}{c}{\textbf{CN}}
& \multicolumn{2}{c}{\textbf{MCI}}
& \multicolumn{2}{c}{\textbf{AD}} \\
\cmidrule(r){2-3}
\cmidrule(r){4-5}
\cmidrule(r){6-7}
& \textbf{No.} & \textbf{Age}
& \textbf{No.} & \textbf{Age}
& \textbf{No.} & \textbf{Age} \\
\midrule
\textbf{Male}   & 37 & 67.41 $\pm$ 7.75 & 2 & 81.00 $\pm$ 7.07 & 1 & 81.00 $\pm$ 0.00 \\
\textbf{Female} & 57 & 68.77 $\pm$ 7.66 & 3 & 67.67 $\pm$ 11.24 & 0 & -- \\
\midrule
\textbf{Total}  & 94 & 68.23 $\pm$ 7.69 & 5 & 73.00 $\pm$ 11.36 & 1 & 81.00 $\pm$ 0.00 \\
\bottomrule
\end{tabular}

\end{table}

\subsection{Cognitive scores} For ADNI, we consider eight scores drawn from standard AD examinations that quantify distinct cognitive domains. We chose these outcomes partly for their clinical standardization and partly to reflect a real-world scenario in which some outcomes are strongly correlated while others are not, allowing us to assess our method on general cases. The scores are: Clinical Dementia Rating Sum of Boxes (CDRSB)~\cite{CDRSB}; the 11- and 13-item Alzheimer's Disease Assessment Scale--Cognitive Subscale (ADAS11, ADAS13) and ADAS Delayed Word Recall (ADASQ4)~\cite{ADAS-cog}; the Mini-Mental State Examination (MMSE)~\cite{MMSE}; and three Rey Auditory Verbal Learning Test (RAVLT)~\cite{RAVLT} scores (immediate recall, learning, and percent forgetting). CDRSB rates impairment across six domains: memory, orientation, judgment and problem-solving, community affairs, home and hobbies, and personal care. ADAS11 measures deficits in memory, language, and praxis through tasks such as word recall and object naming, and ADAS13 extends it with additional tests of memory and executive function, while ADASQ4 assesses episodic memory. MMSE screens for cognitive impairment via concentration, attention, language, and visuospatial skills. Among the RAVLT scores, immediate recall evaluates episodic memory and learning, learning quantifies learning efficiency, and percent forgetting captures delayed memory as the information not retained after a delay. For preliminary external feasibility validation, we used the cognitive measures available in both cohorts. The Clinical Dementia Rating Sum of Boxes is denoted as CDRSB in ADNI and CDRSUM in OASIS3~\cite{CDRSBandCDRSUM}; therefore, these variables were treated as the same cognitive measure. MMSE was used directly, as it is defined identically in both datasets.

\subsection{Image preprocessing} 
\begin{figure*}[!t]
    \begin{center}
    \includegraphics[width=\linewidth]{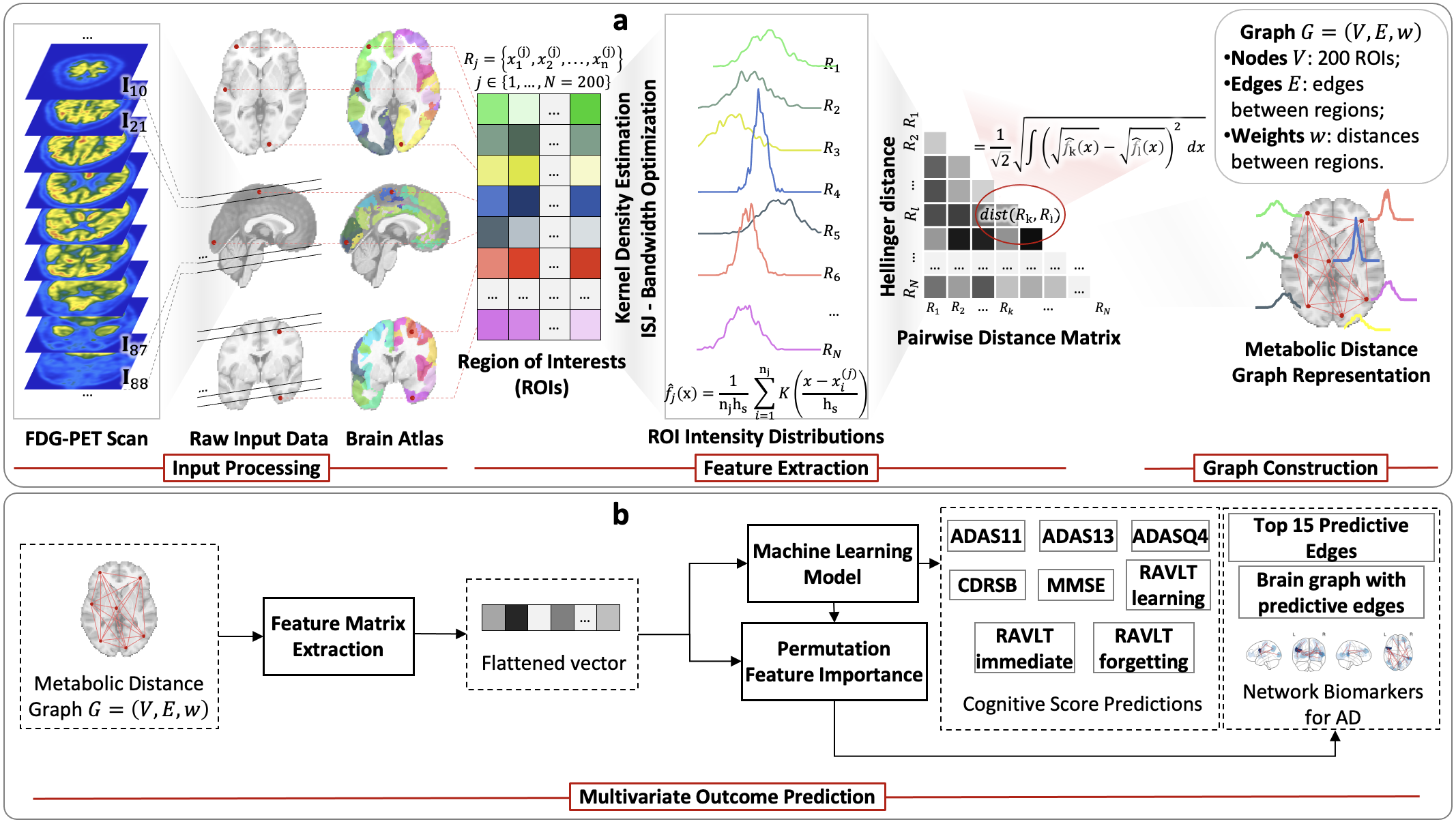}
    \caption{\textbf{A schematic representation of the explainable graph-theoretical machine learning (XGML) framework.} (a) Construction of metabolic distance graphs. (b) From high-dimensional brain graphs to multivariate outcome prediction.}
    \label{fig:model}
    \end{center}
    \vspace{-0.5cm}
\end{figure*}

We utilized FDG-PET scans from both the ADNI and OASIS3 cohorts. The images were preprocessed using the PETSurfer pipeline~\cite{greve2014pet,greve2016petsurfer}, which relies on FreeSurfer-based structural processing to co-register PET with each subject's T1-weighted MRI in native anatomical space. Briefly, for each subject, the individual PET frames were first motion-corrected and then averaged across the temporal dimension to obtain a single representative 3D mean FDG-PET volume. The mean PET image was subsequently co-registered to the subject's T1-weighted MRI in native anatomical space, enabling accurate alignment of metabolic and structural information. Next, PET intensities were normalized using the pons as a reference region to generate standardized uptake value ratio (SUVR) maps. Finally, the Schaefer 2018 atlas with 200 ROIs~\cite{Schaefer2018} was projected onto each subject's cortical surface, allowing us to extract parcel-wise voxel-level FDG-PET SUVR values for graph construction (see Fig.~\ref{fig:model}a).

\subsection{Encoding subject-specific neuroimaging data}
We now define distance graphs. First, to capture the underlying distribution
of voxel intensities in each ROI, we employ KDE.
More specifically, let \( R_j = \{x_1^{(j)}, x_2^{(j)}, \ldots, x_{n_j}^{(j)}\} \)
be the set of FDG-PET voxel SUVR values for a given ROI $R_j$, for
$j\in \{1,\ldots,200\}$, assumed to be drawn from an unknown distribution with
an underlying density \( f_j(x) \). We estimate this density using KDE as:
\[
\hat{f}_j(x) = \frac{1}{n_j h_s} \sum_{i=1}^{n_j}
K\!\Bigl(\frac{x - x_i^{(j)}}{h_s}\Bigr),
\]
where \(K(\cdot)\) is a kernel function satisfying
\(\int_{-\infty}^{+\infty} K(u)\,du = 1\), and \(h_s > 0\) is a subject-specific bandwidth shared across all ROIs of the corresponding subject. For further details on KDE, we refer the reader to~\cite{KDE}. Since the bandwidth generally has a greater influence on the estimate than the choice of kernel~\cite{MARRON1988195}, we use the standard
Gaussian kernel and set $h_s$ with the Improved Sheather-Jones (ISJ) method~\cite{10.1214/10-AOS799}, taking the median of the per-ROI ISJ bandwidths as a single
per-subject value, and resorting to a robust fallback when too few of the
per-ROI ISJ estimates are valid. All of a subject's ROI densities are then
evaluated on a common grid so that the regions can be compared directly,
yielding one probability density function per brain region.

We then quantify the distance between the estimated probability density functions (PDFs) of each pair of ROIs using the
Hellinger distance. The Hellinger distance measures the discrepancy between two
probability densities through the overlap of their square roots; it is symmetric and bounded in $[0,1]$, taking the value 0 when the two densities are identical and 1 when their supports do not overlap. For two ROIs $R_k$ and $R_{\ell}$ with estimated densities
$\hat{f}_k$ and $\hat{f}_{\ell}$, the Hellinger
distance is defined as:
\[
\mathrm{dist}(R_k,R_{\ell}) \;=\; \frac{1}{\sqrt{2}}
\sqrt{
    \int
    \left(
        \sqrt{\hat{f}_k(x)} - \sqrt{\hat{f}_{\ell}(x)}
    \right)^2
    \, dx
}.
\]
In our implementation, the integral is evaluated on the common subject-specific grid using trapezoidal quadrature. Note that this distance cannot be negative, since the integrand is a
squared difference and the square root is taken of a non-negative quantity. For
more details on the Hellinger distance, we refer the reader to~\cite{gibbs_su}. 

To assess the effect of the chosen edge weight, we additionally constructed
graphs using five alternative representations. The continuous $L_2$ distance,
$\bigl(\int(\hat{f}_k-\hat{f}_{\ell})^2\,dx\bigr)^{1/2}$, is computed on the same
KDE densities using the same trapezoidal quadrature. We also included a KDE+DTW graph representation, in which the discretized ROI density curves are compared using Dynamic Time Warping (DTW)~\cite{DTW}.
The remaining three weights are computed
from the raw ROI voxel intensities: the Wasserstein-1
distance~\cite{wasserstein_ref} $\int|F_k(x)-F_{\ell}(x)|\,dx$, where $F_k$ and
$F_{\ell}$ are the empirical cumulative distribution functions; the energy
distance~\cite{szekely_energy}
$\bigl(2\,\mathbb{E}|X-Y|-\mathbb{E}|X-X'|-\mathbb{E}|Y-Y'|\bigr)^{1/2}$, where
$X,X'$ are independent draws from the intensity distribution of $R_k$ and
$Y,Y'$ from that of $R_{\ell}$; and a mean-difference weight
$|\bar{x}_k-\bar{x}_{\ell}|$, the absolute difference of the two ROI mean
intensities.

For each of the evaluated representations, we construct the \emph{distance graph}, which is an undirected, weighted graph \(G = (V, E, w)\), where \(V\) is the set of vertices, corresponding to the 200 ROIs; \( E \) is the set of all edges connecting pairs of distinct vertices; and \( w: E \rightarrow \mathbb{R}_{\geq 0} \) is a weight function assigning a non-negative real number to each edge. More specifically, the weight of an edge \( uv \in E \) is given by the distance between the corresponding ROIs using the respective distance measure (DTW, \(L_2\), Hellinger, Wasserstein, energy distance, or mean difference).

Each representation therefore yields a complete graph with $\binom{200}{2}=19{,}900$ weighted edges per subject, whose weights serve as the input features for the learning models described in the next subsection.

\subsection{Explainable graph-theoretical machine learning}
In what follows, we define the XGML framework, illustrated in
Fig.~\ref{fig:model}. We first benchmarked the six graph representations
described above in combination with six machine-learning models: linear regression, ridge regression, multitask elastic net, random forest, a multilayer perceptron (MLP), and kernel support vector regression (kernel SVR) with a radial basis function (RBF) kernel. Among the evaluated configurations, the KDE+Hellinger
representation combined with random forest achieved the highest predictive
performance (Section~\ref{sec:resultsanddiscussions}). We therefore use this
configuration as the main XGML model in the subsequent analysis.

For each subject, the selected KDE+Hellinger graph is represented by its $200\times200$ weighted adjacency matrix. Because the graph is undirected, the upper-triangular elements excluding the diagonal are flattened into a 19,900-dimensional feature vector used to predict the eight cognitive scores.

For internal cross-subject validation on the ADNI cohort, we employ a repeated
stratified three-fold cross-validation scheme with ten repetitions, giving 30
outer folds in total. The outer folds are stratified by diagnostic group (CN, MCI, and AD). Within each outer training split, model hyperparameters are optimized using Optuna with five-fold stratified inner cross-validation, using only the outer training data to avoid
information leakage. The tuning objective is the variance-normalized mean
squared error averaged across the eight outcomes, and the targets are
standardized within a \texttt{TransformedTargetRegressor} during model fitting.

Linear regression is fitted without hyperparameter tuning, whereas the remaining models are tuned using 25 Optuna trials (30 for kernel SVR). For kernel SVR, the RBF kernel parameter is defined as
$\gamma=\alpha/p$, where $p$ is the feature dimensionality and $\alpha$ is
optimized over $[10^{-2},10^{2}]$ on a logarithmic scale.

To assess whether the graph representations provide predictive information beyond regional summaries of the same FDG-PET data, we additionally evaluate three non-graph baselines: 200 ROI mean intensities; 800 ROI-level distributional features ($200\times4$: mean, standard deviation, skewness, and excess kurtosis); and the latter augmented with age and sex (802 features). All baselines are evaluated using the same nested cross-validation framework.

Model performance is assessed using the Pearson correlation coefficient ($r$),
root mean squared error (RMSE), mean absolute error (MAE), and coefficient of
determination ($R^2$). Within each repetition, predictions from the three outer
test folds are concatenated so that every subject contributes one out-of-fold
prediction, and each metric is computed separately for each cognitive outcome.
For $r$, the ten repetition-specific correlations are averaged using Fisher-z-transformation. An overall correlation across the eight outcomes is then
obtained using an equally weighted Fisher-z-average.

To identify the network edges contributing most strongly to the predictions,
we perform a repeated nested held-out permutation feature importance analysis
of the selected XGML model, i.e., the KDE+Hellinger representation combined
with the joint eight-output random forest. The analysis uses five balanced
subsamples, each comprising 150 subjects (50 CN, 50 MCI, and 50 AD), generated
independently using predefined random seeds (11, 22, 33, 44, and 55); therefore,
subjects may occur in more than one subsample. For each subsample, five-fold
stratified outer cross-validation is performed. Within each outer training
partition, random forest hyperparameters are optimized using 25 Optuna trials
and five-fold stratified inner cross-validation. The optimized model is then
refitted on the complete outer training partition, and permutation importance
is evaluated exclusively on the corresponding held-out test fold using five
permutations per edge. For each cognitive outcome, edge importance is defined
as the increase in held-out RMSE after permutation, yielding eight
outcome-specific importance profiles from the same fitted multi-output forest.

Across the 25 held-out evaluations per cognitive outcome, edges with positive permutation importance are ranked within each evaluation and the top 15 were retained. The final top 15 are selected by frequency of occurrence across these lists, with mean and then median held-out permutation importance used to break ties. Positive importance serves only as an eligibility criterion and is not interpreted as a statistical significance threshold.

\section{Results}
\label{sec:resultsanddiscussions}
Here, we present the results of applying XGML to ADNI and OASIS3 data. We first report the internal cross-subject validation results on the ADNI cohort, including comparisons across the evaluated graph representations, machine learning methods, and non-graph baselines. We then present preliminary external feasibility validation obtained by training on ADNI and testing on the independent OASIS3 cohort for the two shared cognitive measures, followed by the permutation importance results highlighting the most predictive brain subgraphs for each cognitive outcome.

\subsection{Analysis of overall distance values}

To investigate differences in the overall magnitude of metabolic distances across diagnostic groups, we first computed the mean distance across all edges of each subject's individual metabolic distance graph. The mean subject-level distance was highest in the MCI group (0.470 $\pm$ 0.108), compared with the CN (0.441 $\pm$ 0.117) and AD (0.439 $\pm$ 0.079) groups. 

\begin{figure}[ht!]
    \centering
    \includegraphics[width=0.65\linewidth]{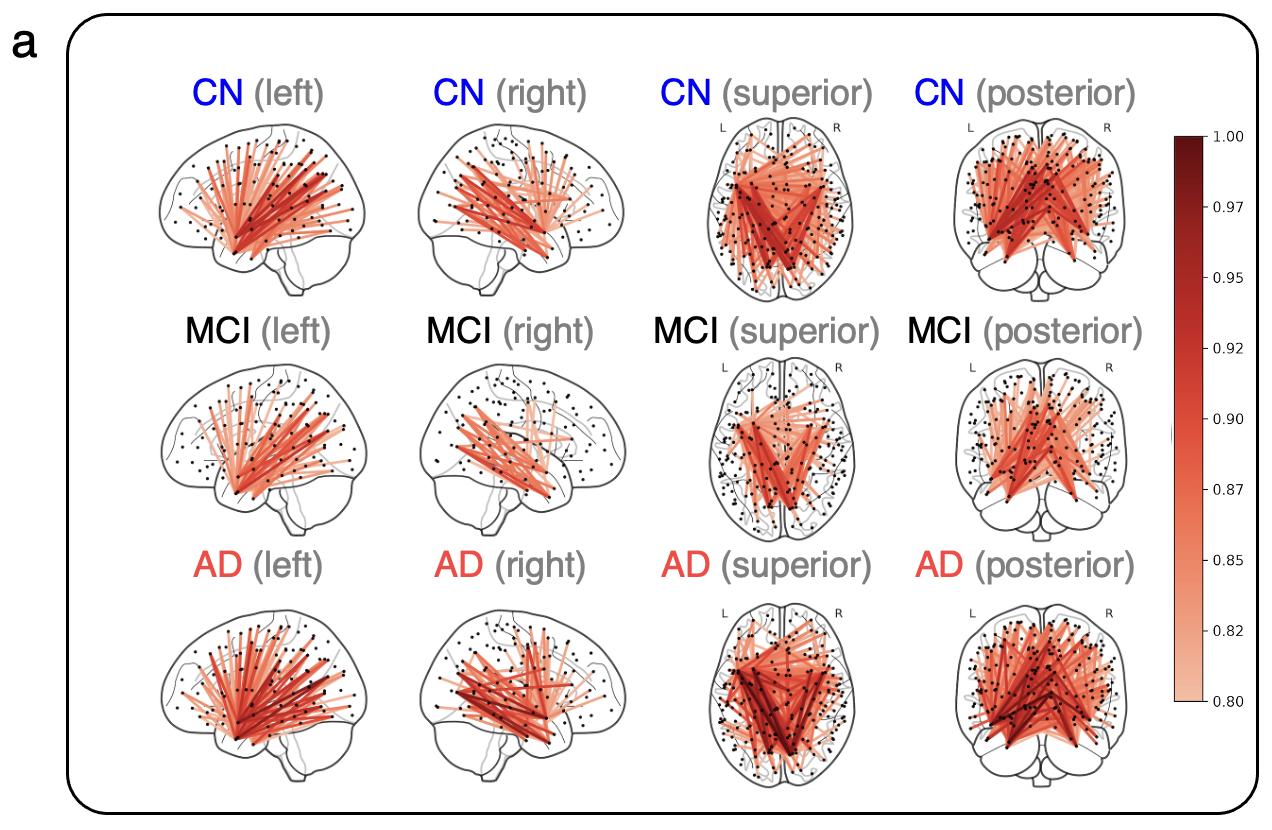}

    \vspace{0.25cm}

    \includegraphics[width=0.82\linewidth]{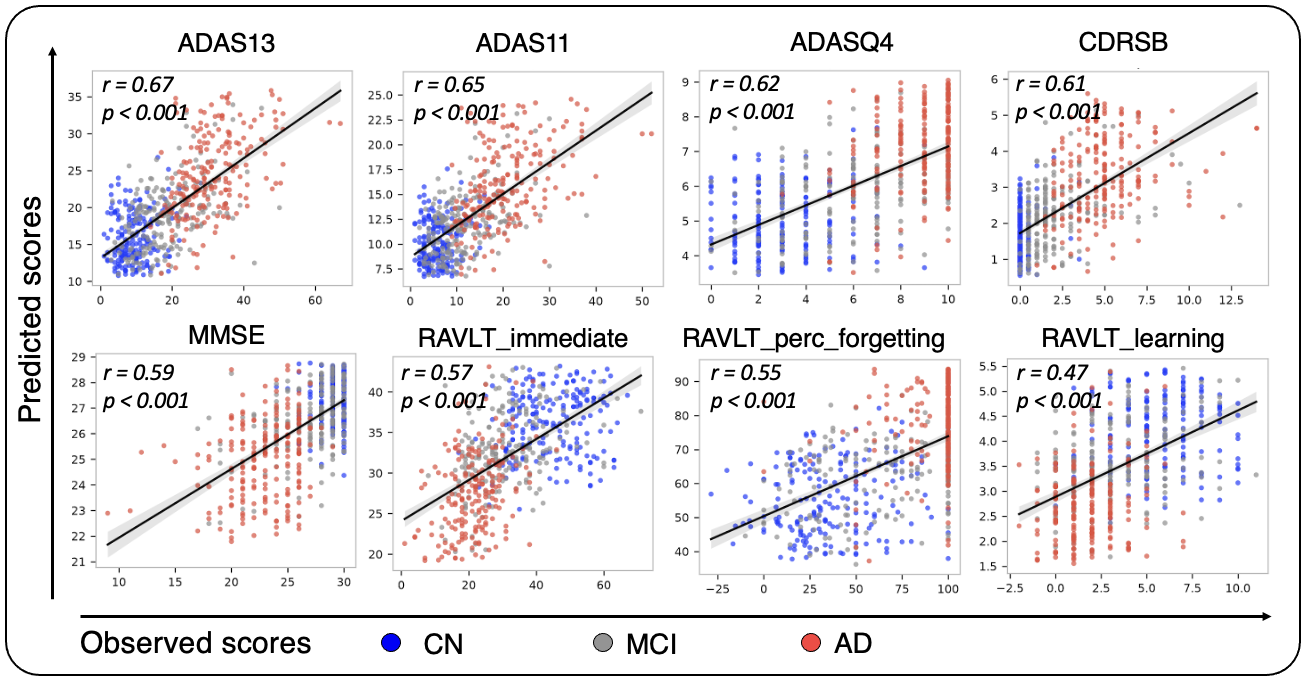}
    \caption{\textbf{Group-level metabolic brain networks and prediction of AD-related cognitive scores.} (a) Group-level FDG-PET metabolic distance networks for CN, MCI, and AD, shown from four viewpoints (left, right, top, and posterior). Edge color indicates normalized metabolic distance, with darker edges representing larger distances. (b) Out-of-sample prediction of eight cognitive scores using KDE+Hellinger graphs and random forest. Observed and predicted values are shown on the x- and y-axes, respectively. Lines show linear fits with 95\% confidence bands; Pearson's $r$ is reported in each panel.}
    \vspace{-0.5cm}
    \label{fig:overall_strength}
\end{figure}

We additionally examined the distribution of distance values in the group-level metabolic graphs. For each diagnostic group, a group-level matrix was obtained by applying a 5\% trimmed mean across the individual distance matrices. The resulting matrices had mean raw distance values of 0.431, 0.463, and 0.432 for CN, MCI, and AD, respectively. Following joint min--max normalization of the three group matrices to $[0,1]$, the corresponding mean values were 0.466, 0.499, and 0.466. The numbers of edges exceeding normalized distance thresholds of 0.50, 0.70, and 0.80 were, respectively, 5810, 1044, and 315 for CN; 8640, 1073, and 169 for MCI; and 5617, 1099, and 380 for AD. For visualization, Fig.~\ref{fig:overall_strength}a shows the group-level metabolic graphs retaining only edges with normalized distance values exceeding 0.80.

\subsection{Predictive performance}

Table~\ref{tab:rep_model_benchmark} summarizes the predictive performance of
the six graph representations and three non-graph baselines across the six
machine learning models for the eight AD-related cognitive outcomes. Overall,
the KDE+Hellinger representation combined with random forest achieved the
highest predictive performance, with an aggregate Fisher-z-averaged Pearson correlation of
$r=0.595$. KDE+Hellinger also yielded the highest performance for five of the
six learners, indicating that predictive performance depended not only on the
learning algorithm but also on the choice of graph representation. Among the
non-graph baselines, the highest performance was obtained with ROI summary
statistics augmented with demographics using kernel SVR ($r=0.540$), which
remained below the best graph-based configuration.

For the selected KDE+Hellinger random forest model, outcome-specific
correlations ranged from $r=0.467$ to $r=0.674$
(Table~\ref{tab:selected_per_outcome}). The strongest predictions were obtained
for ADAS13 ($r=0.674$), ADAS11 ($r=0.653$), ADASQ4 ($r=0.615$), and CDRSB
($r=0.609$), whereas RAVLT\_learning showed the lowest correlation
($r=0.467$). Fig.~\ref{fig:overall_strength}b shows the corresponding
out-of-sample predictions across the eight cognitive outcomes.

\begin{table}[ht!]
\centering
\caption{\textbf{Representation $\times$ model benchmark.} Predictive performance
for six graph representations and three non-graph baselines across six learners.
Each entry reports the equal-weight Fisher-z average across the eight outcome-specific Pearson correlations, where each outcome correlation is Fisher-z averaged across the 10 cross-validation repetitions. For each learner, the highest value is shown in bold. ROI sm denotes ROI summary statistics.}
\label{tab:rep_model_benchmark}
\vspace{0.3em}

\resizebox{\linewidth}{!}{%
\begin{tabular}{lcccccc}
\toprule
Representation & Linear regression & Ridge regression & Multitask elastic net & Random forest & Multilayer perceptron & Kernel SVR \\
\midrule
KDE + DTW & 0.038 & 0.113 & 0.229 & 0.509 & 0.194 & 0.458 \\
KDE + $L_2$ & 0.457 & 0.539 & 0.540 & 0.583 & 0.434 & 0.573 \\
KDE + Hellinger & \textbf{0.493} & \textbf{0.565} & \textbf{0.576} &
\textbf{0.595} & 0.428 & \textbf{0.575} \\
Wasserstein & 0.374 & 0.512 & 0.521 & 0.573 & 0.445 & 0.568 \\
Energy & 0.434 & 0.536 & 0.545 & 0.581 & 0.439 & 0.571 \\
Mean-difference & 0.378 & 0.508 & 0.517 & 0.569 & 0.449 & 0.567 \\
\midrule
ROI mean & 0.221 & 0.462 & 0.451 & 0.511 & 0.462 & 0.515 \\
ROI sm & 0.157 & 0.429 & 0.389 & 0.507 & 0.434 & 0.539 \\
ROI sm + demographics & 0.157 & 0.432 & 0.393 & 0.506 &
\textbf{0.464} & 0.540 \\
\bottomrule
\end{tabular}%
}
\end{table}

\begin{table}[ht!]
\centering
\caption{\textbf{Per-outcome prediction performance of the selected model (KDE+Hellinger with random forest).} For each cognitive outcome, $r$ is the Fisher-z-averaged Pearson correlation, and $\pm$ denotes the standard deviation of the 10 repetition-specific correlations. RMSE, MAE, and $R^2$ are averaged across the 10 repetitions, with RMSE and MAE reported on the original scale of each cognitive outcome.}
\label{tab:selected_per_outcome}
\vspace{0.3em}

\setlength{\tabcolsep}{5pt}
\begin{tabular}{lcccc}
\toprule
Cognitive outcome & $r$ & RMSE & MAE & $R^2$ \\
\midrule
CDRSB & 0.609\,$\pm$\,0.007 & 2.096 & 1.578 & 0.347 \\
ADAS11 & 0.653\,$\pm$\,0.008 & 6.486 & 4.936 & 0.399 \\
ADAS13 & 0.674\,$\pm$\,0.006 & 8.809 & 6.916 & 0.425 \\
ADASQ4 & 0.615\,$\pm$\,0.006 & 2.464 & 2.044 & 0.354 \\
MMSE & 0.593\,$\pm$\,0.009 & 2.934 & 2.296 & 0.332 \\
RAVLT\_immediate & 0.570\,$\pm$\,0.010 & 10.935 & 8.574 & 0.309 \\
RAVLT\_learning & 0.467\,$\pm$\,0.005 & 2.332 & 1.871 & 0.209 \\
RAVLT\_perc\_forgetting & 0.554\,$\pm$\,0.010 & 28.862 & 24.271 & 0.290 \\
\bottomrule
\end{tabular}
\end{table}

\subsection{Preliminary external feasibility validation on OASIS3}

\begin{figure}[h!]
    \centering
    \includegraphics[width=0.75\linewidth]{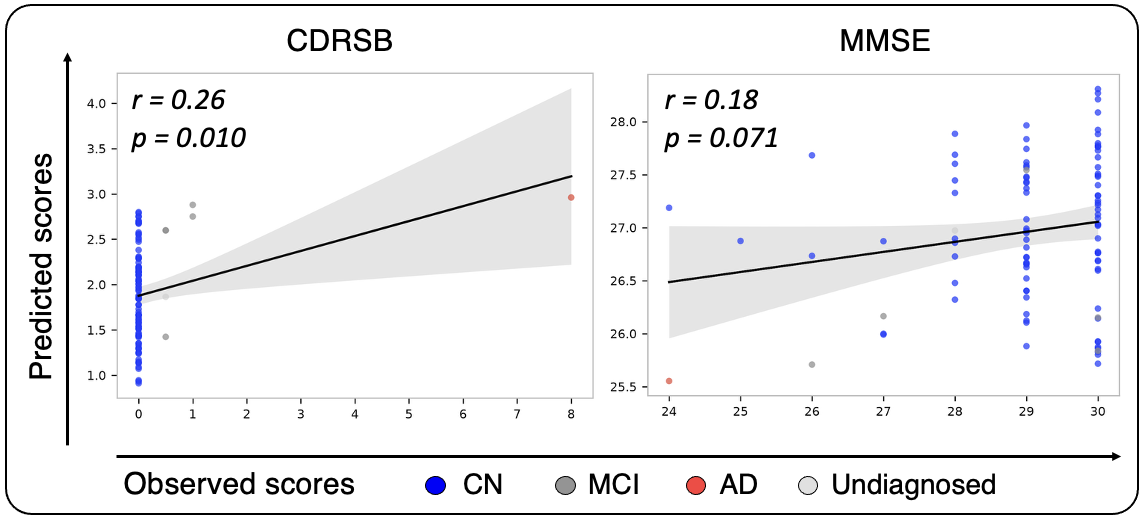}
    \caption{\textbf{Preliminary external feasibility validation on OASIS3.}
    Observed versus predicted CDRSB and MMSE scores for OASIS3 subjects
    using the KDE+Hellinger random forest model trained on ADNI. Each point
    represents one subject and is colored by diagnostic group.}
    \vspace{-0.5cm}
    \label{fig:oasis_external_validation}
\end{figure}

Here, we present the results of the preliminary external feasibility validation of the selected KDE+Hellinger random forest model on the independent OASIS3~\cite{oasis3} cohort. The model was trained on ADNI and evaluated on OASIS3 subjects. Differences in FDG-PET feature distributions were observed between ADNI and OASIS3, consistent with differences in data acquisition between the two cohorts. The raw KDE+Hellinger edge weights were used as input features without additional feature normalization, consistent with the internal ADNI analysis. The model achieved a Pearson correlation of $r=0.255$ ($p=0.010$) for CDRSB and $r=0.181$ ($p=0.071$) for MMSE (Fig.~\ref{fig:oasis_external_validation}).

The OASIS3 cohort was strongly imbalanced, comprising 94 CN, 5 MCI, 1 AD, and 1 subject without a diagnostic label. Cognitive-score variability was also limited: 94 of the 101 subjects had a CDRSB score of zero, while MMSE scores were concentrated near the upper end of the scale (mean $29.05 \pm 1.29$, range 24--30). Given the small and highly imbalanced external cohort, the limited variability in cognitive scores, and the differences in data acquisition between ADNI and OASIS3, these findings should be interpreted as preliminary evidence of external feasibility.

\subsection{Top fifteen predictive network features and their neurological insights}

To assess the reproducibility of the identified top 15 most important edges, we examined their selection
frequency, positive-importance frequency, and pairwise overlap between
evaluation-specific top-15 edge sets (Table~\ref{tab:edge_stability}). Across
cognitive scores, the most frequently selected edge appeared in 24--28\% of the
25 held-out evaluations, while the median selection frequency among the final
top-15 edges ranged from 12--20\%. The selected edges exhibited positive permutation importance in 48--56\% of the evaluations, and mean Jaccard similarities ranged from 0.016 to 0.032.

\begin{figure}
    \centering
    \includegraphics[width=0.8\linewidth]{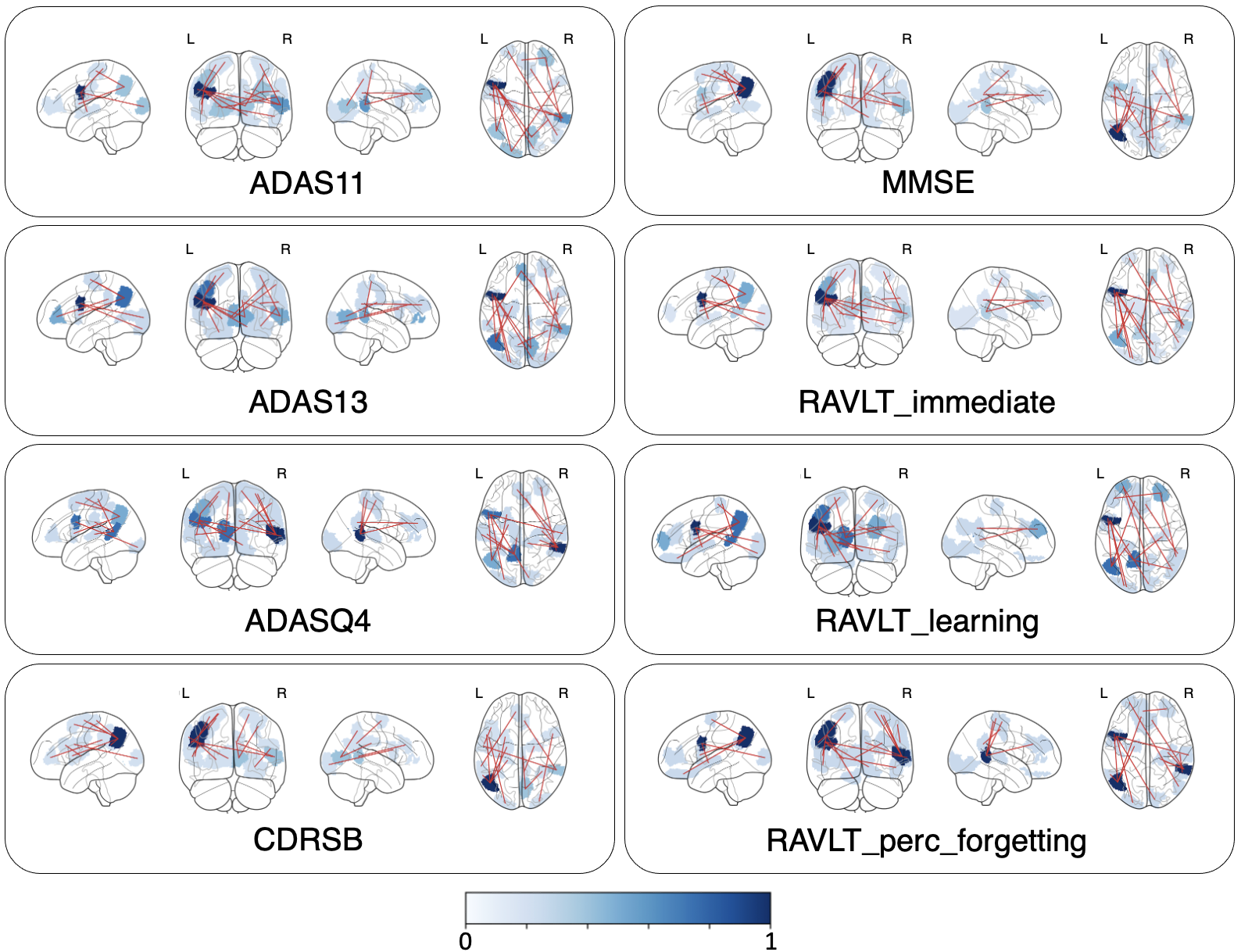}
    \caption{\textbf{Top 15 most predictive edges for each cognitive score.} Each box corresponds to a cognitive score and shows the most important edges, from four viewpoints: left, posterior, right, and top, for predicting that score. The color gradient of the brain regions reflects their normalized involvement in the selected predictive subgraph, quantified as the number of top predictive edges incident to each region, with darker blue indicating involvement in a larger number of selected edges.}
    \label{fig:most_pred_edges}
\end{figure}

To evaluate the contribution each brain region makes to the prediction, we counted the number of top predictive edges incident to each brain region and min–max normalized these counts within every cognitive score. Regions connected to a larger number of the selected predictive edges were considered more relevant for the corresponding cognitive score (see Fig.~\ref{fig:most_pred_edges}).

\begin{table}[ht!]
\centering
\caption{\textbf{Reproducibility of the selected predictive edges across the 25 held-out evaluations.} For each cognitive score, we report the maximum and median selection frequency among the final top 15 edges, the mean pairwise Jaccard similarity between evaluation-specific top-15 edge sets, and the mean frequency with which the selected edges exhibited positive permutation importance. Frequencies are expressed as proportions of the 25 evaluations. Freq. denotes frequency; pos. denotes positive; imp. denotes importance.}
\label{tab:edge_stability}
\vspace{0.3em}

\setlength{\tabcolsep}{3pt}
\begin{tabular}{lcccc}
\toprule
\textbf{Score} & \textbf{Max} & \textbf{Median} & \textbf{Mean} & \textbf{Mean pos.} \\
& \textbf{freq.} & \textbf{freq.} & \textbf{Jaccard} & \textbf{imp. freq.} \\
\midrule
CDRSB & 0.24 & 0.16 & 0.025 & 0.56 \\
ADAS11 & 0.24 & 0.16 & 0.024 & 0.50 \\
ADAS13 & 0.24 & 0.16 & 0.028 & 0.55 \\
ADASQ4 & 0.24 & 0.16 & 0.023 & 0.51 \\
MMSE & 0.28 & 0.20 & 0.032 & 0.54 \\
RAVLT\_immediate & 0.28 & 0.16 & 0.024 & 0.51 \\
RAVLT\_learning & 0.24 & 0.12 & 0.016 & 0.49 \\
RAVLT\_perc\_forgetting & 0.24 & 0.16 & 0.020 & 0.48 \\
\bottomrule
\end{tabular}
\end{table}

The most predictive regions identified for ADAS11, ADAS13, and ADASQ4, which assess memory, language, and praxis-related functions, are predominantly located in the frontal, parietal, and temporal lobes. This observation is consistent with the known neuroanatomical functions of these regions, as memory-related processes involve the temporal, frontal, and parietal lobes, and praxis relies heavily on parietal lobe integrity. The most predictive regions identified for CDRSB, which evaluates orientation, judgment and problem-solving, community affairs, home and hobbies, and personal care, mainly span the frontal, parietal, and temporal lobes. These findings are consistent with the involvement of frontal and parietal regions in orientation and executive functioning, as well as the contribution of parietal areas to sensory integration and activities of daily living. For MMSE, which measures attention, concentration, memory, naming, and visuospatial abilities, the most predictive regions are similarly distributed across the frontal, temporal, and parietal lobes. These regions are well known to support the broad range of cognitive processes assessed by MMSE. Finally, the RAVLT scores capture distinct memory-related functions: RAVLT\_immediate evaluates immediate episodic memory, RAVLT\_learning assesses learning efficiency, and RAVLT\_perc\_forgetting measures delayed memory retention. The predictive regions identified by XGML are primarily located within the frontal, temporal, and parietal lobes. This observation is in agreement with the established role of the temporal lobe in memory encoding and retrieval, together with the contribution of frontal regions to executive aspects of memory processing and parietal regions to memory retrieval mechanisms. See \cite{BrainLobes,brainlobes2} for a reference on the functions of the brain lobes discussed herein.

\begin{figure}
    \centering
    \includegraphics[width=0.8\linewidth]{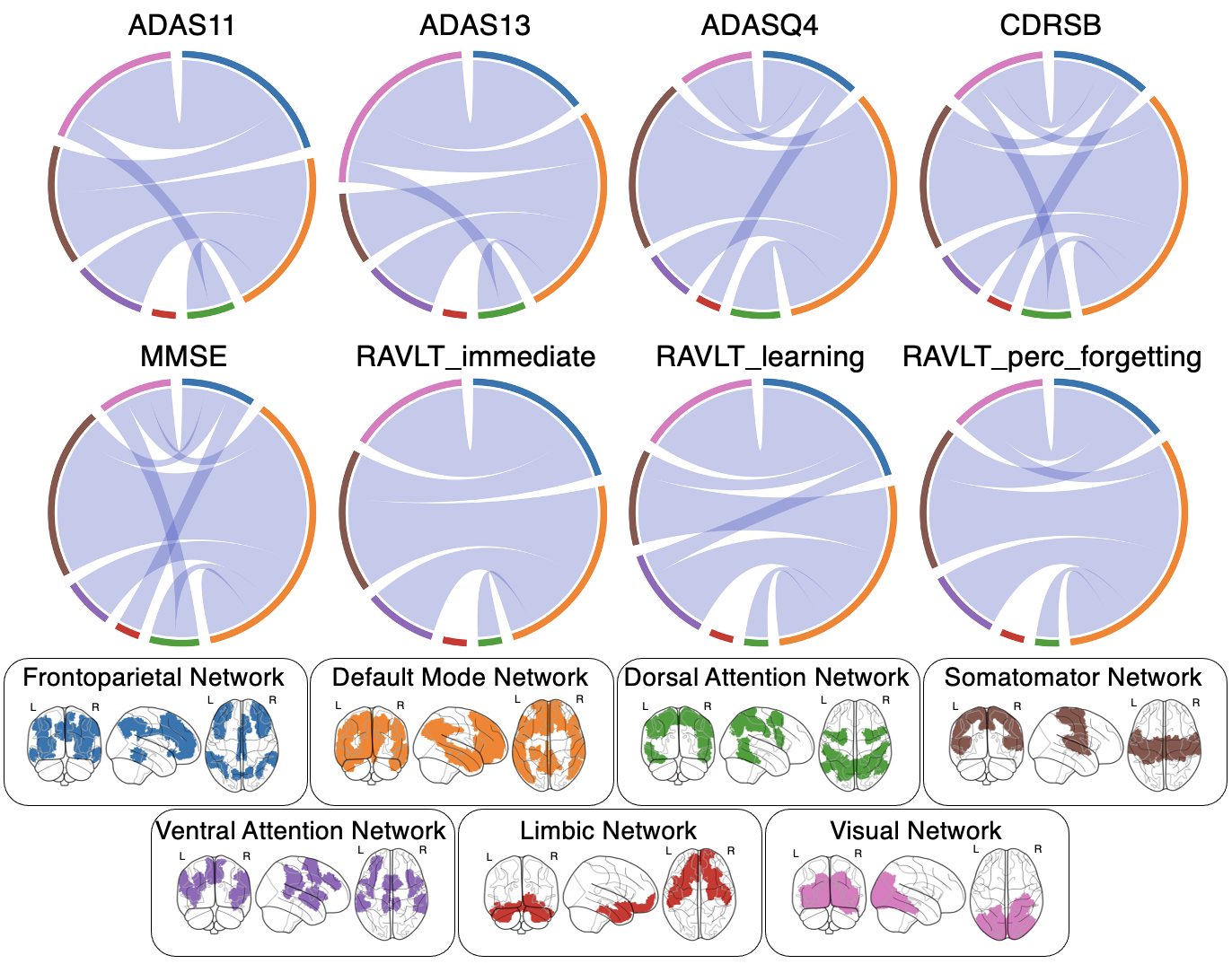}
    \caption{\textbf{The functional distribution of the top 15 most predictive edges for each of the eight cognitive scores.} 
    Each plot is color-coded to match the seven Yeo networks shown in the boxes below.}
    \label{fig:circular}
\end{figure}

We next investigated how the top 15 predictive edges for each cognitive score are functionally distributed in the brain. To do so, we mapped the key edges onto the Yeo 7 functional networks~\cite{Yeo2011}. Across cognitive scores, the predictive edges predominantly involved the Default Mode Network, which is implicated in self-referential processing and memory retrieval~\cite{default}. For ADAS13, ADASQ4, CDRSB, MMSE, and RAVLT\_perc\_forgetting, the top predictive edges mainly had endpoints within this network. For ADAS11, RAVLT\_immediate, and RAVLT\_learning, the predictive edges predominantly involved both the Default Mode and Frontoparietal networks, the latter supporting higher-order cognitive control and working memory~\cite{frontoparietal}. See Fig.~\ref{fig:circular} for details.

Furthermore, we compared the average metabolic distance values between CN and AD
subjects for the top 15 predictive edges identified for each cognitive score
(Table~\ref{tab:cn_ad_top15}). For each selected edge, the mean KDE+Hellinger
distance was computed separately across the CN and AD groups. For six of the eight cognitive scores, the majority of the selected edges showed higher average
distance values in AD than in CN: 11 of 15 edges for CDRSB and ADAS13, 10 of 15
edges for MMSE, and 9 of 15 edges for ADAS11, RAVLT\_immediate, and RAVLT\_perc\_forgetting. In contrast, higher average distances in CN were observed for 9 of
15 edges for ADASQ4 and 8 of 15 edges for RAVLT\_learning. The mean signed
difference (AD$-$CN) was positive for seven outcomes, ranging from $0.007$ for
RAVLT\_learning to $0.100$ for ADAS13, and was slightly negative for ADASQ4
($-0.009$). Overall, the selected predictive edges tended to exhibit higher metabolic distances in AD than in CN, although this pattern was not uniform across outcomes or individual edges.  In the present graph representation, a higher metabolic distance indicates greater dissimilarity between the distributions of FDG-PET glucose metabolism in the two connected ROIs. Thus, where present, higher distances in AD indicate greater metabolic differentiation between the regions forming these predictive connections.

\begin{table}[ht!]
\centering
\caption{\textbf{Top 15 predictive edges: AD versus CN comparison.}
For each cognitive score, we report the number of edges with higher mean
metabolic distances in AD versus CN and the mean signed difference (AD$-$CN);
positive and negative values indicate higher distances in AD and CN, respectively.}
\label{tab:cn_ad_top15}
\vspace{0.3em}

\setlength{\tabcolsep}{3.5pt}
\begin{tabular}{lccc}
\toprule
\textbf{Cognitive Score} & \textbf{AD $>$ CN} & \textbf{CN $>$ AD} & \textbf{Mean AD$-$CN} \\
\midrule
CDRSB                         & 11 & 4 & $0.09$ \\
ADAS11                        & 9 & 6 & $0.05$ \\
ADAS13                        & 11 & 4 & $0.10$ \\
ADASQ4                        & 6 & 9 & $-0.01$ \\
MMSE                          & 10 & 5 & $0.07$ \\
RAVLT\_immediate              & 9 & 6 & $0.06$ \\
RAVLT\_learning               & 7 & 8 & $0.01$ \\
RAVLT\_perc\_forgetting       & 9 & 6 & $0.06$ \\
\bottomrule
\end{tabular}
\end{table}

\begin{figure}[ht!]
    \centering
    \includegraphics[width=0.9\linewidth]{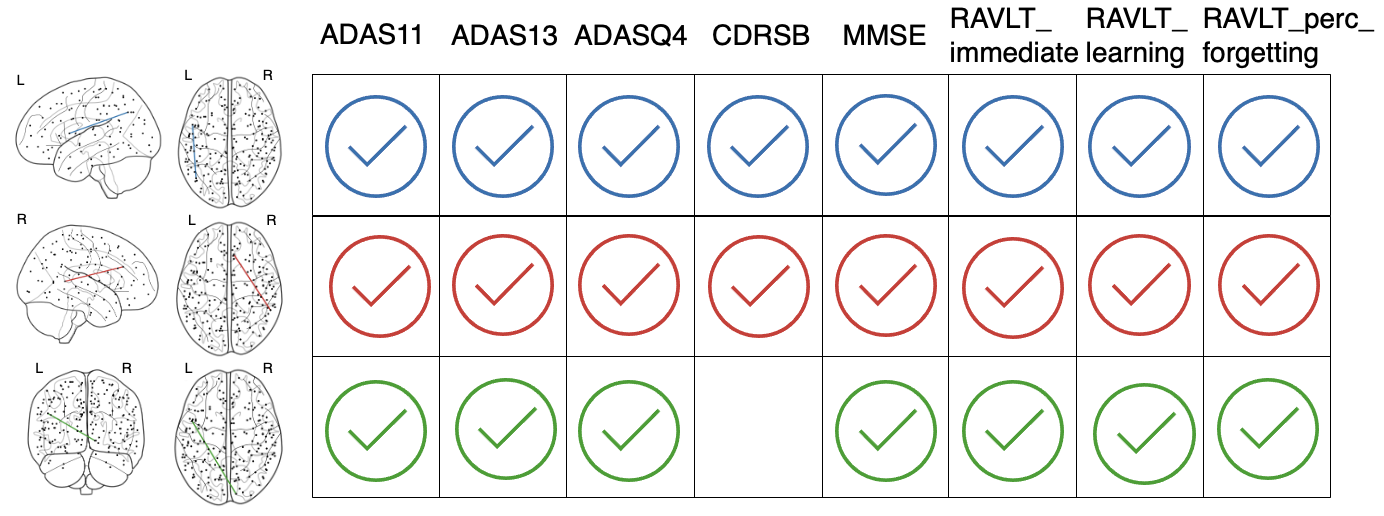}
    \caption{\textbf{Shared top predictive edges across cognitive scores.} The three most frequently occurring predictive edges are shown in the figure. Each row corresponds to a predictive edge, and each column represents a cognitive score. A check mark indicates that the given edge was among the top 15 most predictive edges for that score.}
    \label{fig:freq_edges}
\end{figure}

Finally, we investigated whether some of the top predictive features were shared across multiple cognitive scores. Our results revealed that several metabolic connections were repeatedly identified across multiple cognitive scores (see Fig.~\ref{fig:freq_edges}). In particular, two predictive edges were identified among the top 15 most predictive features for all eight cognitive scores, while one additional edge was shared across seven cognitive scores. Furthermore, three edges were shared across six cognitive scores, two across five cognitive scores, and three across four cognitive scores. Among the remaining predictive connections, nine edges were identified for three cognitive scores, eight for two cognitive scores, and fourteen were specific to a single cognitive score, resulting in a total of forty-two unique predictive edges across all cognitive scores.

\section{Discussion and Conclusion}
\label{sec:discussions}

In this paper, we introduced explainable graph-theoretical machine learning (XGML), a framework that combines insights from graph theory, machine learning, and neurobiology. More specifically, XGML encodes subject-specific neuroimaging data into weighted brain graphs, uses their edge weights as features for multivariate prediction, and identifies subgraphs that contribute most strongly to individual outcomes. To evaluate the framework, we constructed individual metabolic distance brain graphs using FDG-PET scans from the ADNI dataset and used their edge weights as network features to predict multivariate AD-related outcomes. We compared six graph representations and three non-graph baselines across six machine learning models, evaluated the neurobiological relevance of the most predictive connections, and performed a preliminary external evaluation on an independent OASIS3 cohort.

The paper makes several methodological contributions. First, we introduce a new approach that involves both constructing individual metabolic brain graphs from FDG-PET data and using their edges as input features to a predictive model. In particular, this approach enables the construction of subject-specific graphs, rather than relying on group-level analyses. Additionally, since the (dis)connections between brain regions in AD are not yet fully understood, we use all edge weights in our analysis instead of applying thresholds, thus avoiding, for example, preselecting connections, as done in previous studies. Second, we compare multiple approaches for constructing individual metabolic brain graphs and benchmark them against non-graph representations based on regional FDG-PET summaries. Our results show that graph-based representations can provide predictive information beyond these simpler regional features, while also demonstrating that the choice of graph construction substantially influences predictive performance. Third, compared to univariate prediction tasks (e.g., assessing AD status or severity) using network features, our framework predicts multivariate AD-related outcomes using brain graphs. Fourth, the framework is explainable: it identifies the brain subgraphs that contribute most to the predictions.

In the literature, graphs constructed from FDG-PET mainly rely on group-level analyses since each scan contains a single numerical value per voxel. When considering brain regions of interest (ROIs), the number of voxels per region varies, making standard statistical methods such as correlation, which require equal measures, impractical. To address this, a typical approach in the literature is to extract a summary statistic (e.g., the mean or median FDG uptake) for each ROI, reducing each region to a single value. Since correlations cannot be computed from a single value, they are calculated across patients, thereby masking subject-specific differences.

In our study, we address the varying number of voxels across ROIs by representing regional FDG-PET information in several ways that naturally accommodate unequal numbers of voxel observations. Among the evaluated graph representations, KDE+Hellinger provided the strongest overall performance and achieved the highest correlation for five of the six learning models. Combined with random forest, it yielded the best-performing representation–model configuration overall. Furthermore, the results of the comparison with the non-graph baselines suggest that the graph-based representations can capture predictive information beyond simpler regional summaries of the same FDG-PET data, with the KDE+Hellinger graph representation outperforming all evaluated non-graph baselines. Together, these results indicate that retaining distributional information within ROIs can be beneficial and suggest that predictive performance depends substantially on how the relationships between distributional information within ROIs are encoded. We would also like to mention that, while graph neural networks (GNNs) have become a powerful approach for graph-structured data, their application in the present setting would require a different modeling formulation. In our framework, each subject is represented by a fully connected graph with identical topology across subjects, and the primary subject-specific information is encoded in the edge weights. Although GNN architectures can incorporate edge information, standard message-passing approaches typically also rely on node representations and additional architectural choices for integrating node and edge features. Adapting such models to the present distance-based representation would therefore constitute a distinct modeling framework and is beyond the scope of this study. Moreover, although we use 200 brain regions throughout the paper, XGML can be generalized to different numbers of ROIs. Finally, while the present study evaluates XGML using FDG-PET data, the framework is not restricted to this imaging modality and could, in principle, be extended to other neuroimaging modalities, such as structural MRI, functional MRI, and amyloid-PET, for constructing subject-specific graphs and identifying predictive subgraphs.

Our group-level analyses (Fig.~\ref{fig:overall_strength}a) revealed differences in the distribution of metabolic distances across diagnostic groups. The MCI group showed the highest mean metabolic distance, both at the subject level and in the group-level graphs, while the number of edges exceeding a given normalized distance threshold varied across groups and depended on the threshold considered. These findings suggest that alterations in metabolic relationships between brain regions across diagnostic groups may be more complex than a uniform increase or decrease in metabolic distance with disease severity. Such differences are broadly consistent with the characterization of AD as a disorder involving altered brain network organization, often referred to as a “disconnection syndrome” \cite{Delbeuck2003}.

Neurobiologically, our findings are broadly consistent with previous reports of altered brain-network organization in AD~\cite{Delbeuck2003} and provide additional insights into the predictive relevance of specific metabolic connections. Predictive performance varied across cognitive scores, with the highest correlations observed for ADAS13, ADAS11, and ADASQ4, followed by CDRSB and MMSE, while lower performance was observed for the three RAVLT measures. These differences suggest that the metabolic graph representation captures information relevant to multiple aspects of cognitive impairment, although the extent to which individual cognitive scores can be predicted from FDG-PET connectivity varies. In addition, the brain connections identified as most predictive for each cognitive score may represent candidate network biomarkers for domain-specific cognitive impairment, although further validation is required. This interpretation is supported by the observation that, for each cognitive score, the most predictive brain regions correspond to regions whose known functions are closely related to the cognitive domains assessed by that score. For example, CDRSB evaluates impairments in memory, orientation, judgment and problem-solving, community affairs, home and hobbies, and personal care, while the most predictive regions identified for CDRSB are predominantly located within the frontal, parietal, and temporal lobes, which are known to support these functions. Similar neuroanatomical correspondence was observed for the remaining cognitive scores. This agreement between the identified predictive regions and their established functional roles supports the neurobiological plausibility of the identified predictive connections and suggests that they warrant further investigation as candidate network biomarkers.

In parallel, we identified several brain connections that were repeatedly selected among the most predictive features across multiple cognitive outcomes. Notably, two edges were shared across all eight outcomes, while additional edges were shared across seven, six, and five outcomes. This pattern suggests that the predictive metabolic connectivity underlying cognitive performance comprises both shared and outcome-specific components. Edges recurring across multiple outcomes may reflect metabolic connectivity patterns related to cognitive processes common to several domains, whereas edges specific to individual or fewer outcomes may capture more distinct aspects of cognitive function. The recurrent edges, particularly those shared across most or all outcomes, therefore warrant further investigation as candidate network biomarkers of cognitive impairment in AD.

As for the potential clinical relevance of XGML, the framework shifts from traditional group-level FDG-PET analysis to the construction of individual metabolic brain graphs. Using the XGML pipeline, a metabolic distance graph can be generated for each subject and subsequently used as input to predictive models of multiple AD-related cognitive scores. The framework also enables the characterization of subject-specific metabolic connectivity patterns, which may provide complementary information to conventional neuroimaging analyses. Our comparison with non-graph regional representations further suggests that these graph-based representations capture predictive information beyond simpler regional FDG-PET summary features. Beyond prediction, XGML identifies brain connections that are most predictive of each cognitive score. These predictive edges may provide insights into the relationship between metabolic brain connectivity and different cognitive domains affected in AD and may represent candidate network biomarkers of cognitive impairment. The recurrence of several predictive edges across multiple cognitive scores further suggests that some metabolic connections may capture shared network-level patterns related to cognitive impairment, while others may be more specific to individual cognitive measures. Future studies are needed to determine whether these network features have potential clinical utility and are reproducible across independent cohorts and longitudinal settings. Finally, because XGML constructs individual metabolic brain graphs, it could potentially be extended to longitudinal neuroimaging studies. Investigating how metabolic connectivity patterns evolve across repeated scans may provide additional insights into disease progression. However, longitudinal data were not considered in the present study, and future work is required to evaluate this application.

As for the computational complexity, let $\mathcal{R}$ denote the number of ROIs, $M$ the total number of voxels across ROIs, and $K_s$ the size of the common KDE evaluation grid for subject $s$. FFT-based KDE evaluation across all ROIs requires $\mathcal{O}(M+\mathcal{R}K_s\log K_s)$ operations~\cite{Gramacki03042017}. Computing the Hellinger distance between all $\mathcal{O}(\mathcal{R}^2)$ pairs of density functions requires $\mathcal{O}(\mathcal{R}^2K_s)$ operations, giving an overall per-subject complexity of
$T=\mathcal{O}\!\left(M+\mathcal{R}K_s\log K_s+\mathcal{R}^2K_s\right)$. We use a base grid of $2048$ points, adaptively enlarged per subject when required to maintain numerical resolution. Graph construction was performed on a Linux server with an AMD Ryzen Threadripper PRO 7965WX processor (24 physical cores, 48 logical cores) and 250\,GB RAM, parallelized across subjects using Python's \texttt{ProcessPoolExecutor} with \texttt{MAX\_WORKERS}=30. For the 600 ADNI subjects, the KDE-based graph-construction run required $13\pm3$ seconds per subject (range 7--22\,s), with grid sizes ranging from 2048 to 22{,}295 points. The timed run generated both KDE+$L_2$ and KDE+Hellinger representations from the same subject-level KDE estimates.

There are several limitations to this study. First, we performed a preliminary external feasibility validation on an independent OASIS3 cohort. However, the model achieved weak correlations for both cognitive scores considered, with the correlation reaching statistical significance for CDRSB but not for MMSE. The predictive performance observed in the ADNI cohort was therefore not replicated in the external dataset. Several factors may have contributed to this result, including differences in imaging acquisition protocols, the highly imbalanced diagnostic composition of the OASIS3 cohort (94 CN, 5 MCI, and 1 AD subject), limited variability in cognitive scores, and the relatively small sample size. Another limitation of the external validation is the temporal alignment between FDG-PET acquisition and cognitive assessment in the OASIS3 cohort. Because the exact dates of cognitive assessments were unavailable, the interval between imaging and cognitive evaluation was approximated using the difference in participant ages. This approximation may have introduced additional temporal uncertainty, potentially weakening the correspondence between imaging-derived metabolic patterns and the associated cognitive measurements and contributing to the reduced predictive performance observed in the external cohort. Consequently, the present external validation should be interpreted cautiously and does not provide sufficient evidence for robust cross-cohort generalizability. Future work should evaluate XGML on larger, more diverse, and better-balanced external datasets to rigorously assess its generalizability. Second, we applied XGML using FDG-PET data, which provides valuable metabolic information but is relatively costly and less accessible due to the use of radiotracers. This may limit the practical deployment of the method in broader clinical settings. Future studies should investigate the applicability of XGML to other neuroimaging modalities, such as structural MRI and functional MRI, as well as multimodal combinations, which may improve both accessibility and predictive performance. Third, the framework involves multiple components, including graph construction and machine learning, each with associated methodological and parameter choices. While we compared several graph representations and machine learning models and performed model tuning within a cross-validation framework, a more exhaustive exploration of graph-construction parameters and alternative representations may further improve predictive performance.

Finally, although the model performed well during internal validation, there remains room for improvement. Extending the framework to longitudinal data may enable modeling of disease progression and improve early prediction, while incorporating complementary modalities, such as genetic, proteomic, or multimodal neuroimaging data, may further enhance predictive accuracy and provide a more comprehensive characterization of AD-related processes~\cite{multimodal,Yuan2012Multi}.

To conclude, explainable graph-theoretical machine learning for identifying brain-network features predictive of multivariate cognitive outcomes remains an emerging area of research. Our findings demonstrate the potential of subject-specific metabolic brain graphs for multivariate prediction and for identifying candidate network biomarkers of cognitive impairment. We hope that the proposed framework encourages further evaluation across independent datasets, neuroimaging modalities, and longitudinal settings, ultimately improving our understanding of how alterations in brain-network organization relate to AD and its cognitive manifestations.

\bigskip
\vspace{3mm} \noindent\textbf{Disclosure of Interests.} The authors have no competing interests to declare that are relevant to the content of this article. 

%
%
%
\newpage
\bibliographystyle{splncs04}
\bibliography{bibliography_v2}

@misc{who_dementia_2023,
  author       = {{World Health Organization}},
  title        = {Dementia},
  year         = {2023},
  howpublished = {\url{https://www.who.int/news-room/fact-sheets/detail/dementia}},
  note         = {Accessed: July 15, 2025}
}

@misc{alzheimers_research_uk_2021,
  author       = {{Alzheimer's Research UK}},
  title        = {Worldwide dementia cases to triple by 2050},
  year         = {2021},
  howpublished = {\url{https://www.alzheimersresearchuk.org/news/worldwide-dementia-cases-to-triple-by-2050/}},
  note         = {Accessed: July 15, 2025}
}

@article{Gramacki03042017,
author = {Artur Gramacki and Jarosław Gramacki},
title = {FFT-Based Fast Computation of Multivariate Kernel Density Estimators With Unconstrained Bandwidth Matrices},
journal = {Journal of Computational and Graphical Statistics},
volume = {26},
number = {2},
pages = {459--462},
year = {2017}
}

@article{greve2014pet,
  title     = {Cortical surface-based analysis reduces bias and variance in kinetic modeling of brain {PET} data},
  author    = {Greve, Douglas N and Svarer, Claus and Fisher, Patrick M and Feng, Ling and Hansen, Adam E and Baare, William and Rosen, Bruce and Fischl, Bruce and Knudsen, Gitte M},
  journal   = {NeuroImage},
  volume    = {92},
  pages     = {225--236},
  year      = {2014},
  publisher = {Elsevier},
  doi       = {10.1016/j.neuroimage.2013.12.021}
}

@article{greve2016petsurfer,
  title     = {Different partial volume correction methods lead to different conclusions: an {$^{18}$F-FDG-PET} study of aging},
  author    = {Greve, Douglas N and Salat, David H and Bowen, Spencer L and Izquierdo-Garcia, David and Schultz, Aaron P and Catana, Ciprian and Becker, J Alex and Buckner, Randy L and Fischl, Bruce and Johnson, Keith A},
  journal   = {NeuroImage},
  volume    = {132},
  pages     = {334--343},
  year      = {2016},
  publisher = {Elsevier}
}

@INPROCEEDINGS{11222021,
  author={Gürbüz, Saruhan Mete and Adel, Mouloud},
  booktitle={2025 Fourteenth International Conference on Image Processing, Theory, Tools \& Applications (IPTA)}, 
  title={An Edge Feature Inclusive Variational Graph Autoencoder for Pet-Driven Alzheimer's Diagnosis}, 
  year={2025},
  volume={},
  number={},
  pages={1-4},
  doi={10.1109/IPTA66025.2025.11222021}}

@article {oasis3,
	author = {LaMontagne, Pamela J. and Benzinger, Tammie LS. and Morris, John C. and Keefe, Sarah and Hornbeck, Russ and Xiong, Chengjie and Grant, Elizabeth and Hassenstab, Jason and Moulder, Krista and Vlassenko, Andrei G. and Raichle, Marcus E. and Cruchaga, Carlos and Marcus, Daniel},
	title = {OASIS-3: Longitudinal Neuroimaging, Clinical, and Cognitive Dataset for Normal Aging and Alzheimer Disease},
	year = {2019},
	doi = {10.1101/2019.12.13.19014902},
	journal = {medRxiv}
}

@article{CDRSBandCDRSUM,
  author  = {O'Bryant, Sid E. and Waring, Stephen C. and Cullum, C. Munro and Hall, James and Lacritz, Laura and Massman, Paul J. and Lupo, Philip J. and Reisch, Joan S. and Doody, Rachelle and Texas Alzheimer's Research Consortium},
  title   = {Staging Dementia Using Clinical Dementia Rating Scale Sum of Boxes Scores: A Texas Alzheimer's Research Consortium Study},
  journal = {Archives of Neurology},
  year    = {2008},
  volume  = {65},
  number  = {8},
  pages   = {1091--1095},
  doi     = {10.1001/archneur.65.8.1091}
}

@article{tuan2025fdg,
  author       = {Tuan, Pham Minh and Adel, Mouloud and Trung, Nguyen Linh and Horowitz, Tatiana and Parlak, Ismail Burak and Guedj, Eric},
  title        = {{FDG-PET}-based brain network analysis: a brief review of metabolic connectivity},
  journal      = {EJNMMI Reports},
  volume       = {9},
  number       = {1},
  pages        = {4},
  year         = {2025},
  publisher    = {Springer}
}

@article{LI2025110579,
title = {Transformer attention-based neural network for cognitive score estimation from sMRI data},
journal = {Computers in Biology and Medicine},
volume = {196},
pages = {110579},
year = {2025},
doi = {10.1016/j.compbiomed.2025.110579},
author = {Songheng Li and Yanteng Zhang and Congyu Zou and Lipei Zhang and Fei Li and Qiang Liu}
}

@article{bhagwat2019ann_alzheimers,
  author       = {Bhagwat, Nikhil and Pipitone, Jon and Voineskos, Aristotle N. and Chakravarty, Mallar M. and Alzheimer's Disease Neuroimaging Initiative},
  title        = {An artificial neural network model for clinical score prediction in Alzheimer disease using structural neuroimaging measures},
  journal      = {Journal of Psychiatry \& Neuroscience},
  year         = {2019},
  volume       = {44},
  number       = {4},
  pages        = {246--260},
  doi          = {10.1503/jpn.180016}
}

@article{snm_alzheimers,
  author       = {Barbara J. Grabher},
  title        = {Effects of Alzheimer Disease on Patients and Their Family},
  journal      = {Journal of Nuclear Medicine Technology},
  volume       = {46},
  number       = {4},
  pages        = {335-340},
  year         = {2018}
}

@article{passeri2022alzheimers,
  author       = {Passeri, Elodie and Elkhoury, Kamil and Morsink, Margaretha and Broersen, Kerensa and Linder, Michel and Tamayol, Ali and Malaplate, Catherine and Yen, Frances T. and Arab-Tehrany, Elmira},
  title        = {Alzheimer's Disease: Treatment Strategies and Their Limitations},
  journal      = {International Journal of Molecular Sciences},
  volume       = {23},
  number       = {22},
  pages        = {13954},
  year         = {2022}
}

@misc{nia_alzheimers_treatment,
  author       = {{National Institute on Aging}},
  title        = {How Alzheimer's Disease is Treated},
  year         = {2023},
  howpublished = {\url{https://www.nia.nih.gov/health/alzheimers-treatment/how-alzheimers-disease-treated}}
}

@article{McKhann2011,
  author    = {McKhann, Guy M. and Knopman, David S. and Chertkow, Howard and others},
  title     = {The diagnosis of dementia due to Alzheimer's disease: Recommendations from the National Institute on Aging-Alzheimer’s Association workgroups on diagnostic guidelines for Alzheimer’s disease},
  journal   = {Alzheimer's \& Dementia},
  volume    = {7},
  number    = {3},
  pages     = {263--269},
  year      = {2011}
}

@article{FDG-PET_most_studies,
  author       = {Huang, Sheng-Yao and Hsu, Jung-Lung and Lin, Kun-Ju and Hsiao, Ing-Tsung},
  title        = {A Novel Individual Metabolic Brain Network for {18F-FDG PET} Imaging},
  journal      = {Frontiers in Neuroscience},
  volume       = {14},
  pages        = {344},
  year         = {2020}
}

@article{FDG_PET,
  author       = {Arbizu, Javier and Morbelli, Silvia and Minoshima, Satoshi and Barthel, Henryk and Kuo, Philip and Van Weehaeghe, Donatienne and Horner, Neil and Colletti, Patrick M. and Guedj, Eric},
  title        = {{SNMMI} Procedure Standard/{EANM} Practice Guideline for Brain [{18F}]{FDG} {PET} Imaging, Version 2.0},
  journal      = {Journal of Nuclear Medicine},
  volume       = {66},
  number       = {Suppl 2},
  pages        = {S45--S60},
  year         = {2025}
}

@article{indivdiff,
title = {Individual differences in cognition, affect, and performance: Behavioral, neuroimaging, and molecular genetic approaches},
journal = {NeuroImage},
volume = {59},
number = {1},
pages = {70-82},
year = {2012},
note = {Neuroergonomics: The human brain in action and at work},
author = {Raja Parasuraman and Yang Jiang}
}

@article{Shokouhi2013,
  author    = {Sepideh Shokouhi and Daniel Claassen and Hakmook Kang and others},
  title     = {Longitudinal progression of cognitive decline correlates with changes in the spatial pattern of brain $^{18}$F-FDG PET},
  journal   = {Journal of Nuclear Medicine},
  year      = {2013},
  volume    = {54},
  number    = {9},
  pages     = {1564--1569}
}

@article{gomez_ramirez2014,
  author    = {Gomez-Ramirez, Jaime and Wu, Jinglong},
  title     = {Network-based biomarkers in Alzheimer’s disease: review and future directions},
  journal   = {Frontiers in Aging Neuroscience},
  volume    = {6},
  pages     = {12},
  year      = {2014}
}

@article{John2017,
  author    = {Majnu John and Toshikazu Ikuta and Janina Ferbinteanu},
  title     = {Graph analysis of structural brain networks in Alzheimer’s disease: beyond small world properties},
  journal   = {Brain Structure and Function},
  volume    = {222},
  pages     = {923--942},
  year      = {2017}
}

@article{Wei2016,
  author    = {Rizhen Wei and Chuhan Li and Noa Fogelson and others},
  title     = {Prediction of Conversion from Mild Cognitive Impairment to Alzheimer's Disease Using MRI and Structural Network Features},
  journal   = {Frontiers in Aging Neuroscience},
  volume    = {8},
  pages     = {76},
  year      = {2016}
}

@article{critedges,
  author       = {Brier, Matthew R. and Thomas, Jewell B. and Ances, Beau M.},
  title        = {Network Dysfunction in Alzheimer's Disease: Refining the Disconnection Hypothesis},
  journal      = {Brain Connectivity},
  volume       = {4},
  number       = {5},
  pages        = {299--311},
  year         = {2014}
}

@article{bassett2017network,
  title={Network neuroscience},
  author={Bassett, Danielle S and Sporns, Olaf},
  journal={Nature Neuroscience},
  volume={20},
  pages={353--364},
  year={2017},
  publisher={Nature Publishing Group}
}

@inproceedings{PET_classification,
author={Guo, Jiaming and Qiu, Wei and Li, Xiang and others},
  booktitle={2019 IEEE International Conference on Big Data (Big Data)}, 
  title={Predicting Alzheimer’s Disease by Hierarchical Graph Convolution from Positron Emission Tomography Imaging}, 
  year={2019},
  volume={},
  number={},
  pages={5359-5363}
}

@article{PET_pred,
title = {Predicting cognitive decline with deep learning of brain metabolism and amyloid imaging},
journal = {Behavioural Brain Research},
volume = {344},
pages = {103-109},
year = {2018},
author = {Hongyoon Choi and Kyong Hwan Jin}
}

@article{Dubois2023,
  title={Biomarkers in Alzheimer’s disease: role in early and differential diagnosis and recognition of atypical variants},
  author={Dubois, Bruno and von Arnim, Christine A.F. and Burnie, Nerida and others},
  journal={Alzheimer's Research \& Therapy},
  volume={15},
  pages ={175},
  year={2023}
}

@article{Delbeuck2003,
  title={Alzheimer's Disease as a Disconnection Syndrome?},
  author={Delbeuck, Xavier and Van der Linden, Martial and Collette, Fabienne},
  journal={Neuropsychology Review},
  volume={13},
  number={2},
  pages={79--92},
  year={2003},
  publisher={Springer}
}

@article{Henkel2020,
  author    = {Rebecca Henkel and Matthias Brendel and Marco Paolini and others},
  title     = {FDG PET Data is Associated with Cognitive Performance in Patients from a Memory Clinic},
  journal   = {Journal of Alzheimer's Disease},
  year      = {2020},
  volume    = {78},
  number    = {1},
  pages     = {207--216}
}

@book{brainlobes2,
     editor    = {Kandel, Eric R. and Koester, John D. and Mack, Sarah H. and Siegelbaum, Steven A.},
     title     = {Principles of Neural Science},
     edition   = {6th},
     publisher = {McGraw Hill},
     address   = {New York, NY},
     year      = {2021}
   }

@article{default,
  author    = {Randy L. Buckner and Jessica R. Andrews-Hanna and Daniel L. Schacter},
  title     = {The brain's default network: anatomy, function, and relevance to disease},
  journal   = {Annals of the New York Academy of Sciences},
  volume    = {1124},
  pages     = {1--38},
  year      = {2008},
  publisher = {New York Academy of Sciences}
}

@article{frontoparietal,
title = {Large-scale brain networks and psychopathology: a unifying triple network model},
journal = {Trends in Cognitive Sciences},
volume = {15},
number = {10},
pages = {483-506},
year = {2011},
author = {Vinod Menon}
}

@article{10.1214/10-AOS799,
author = {Z. I. Botev and J. F. Grotowski and D. P. Kroese},
title = {{Kernel density estimation via diffusion}},
volume = {38},
journal = {The Annals of Statistics},
number = {5},
publisher = {Institute of Mathematical Statistics},
pages = {2916--2957},
year = {2010}
}

@article{Schaefer2018,
  title={Local-global parcellation of the human cerebral cortex from intrinsic functional connectivity MRI},
  author={Schaefer, Alexander and Kong, Ru and Gordon, Evan M and others},
  journal={Cerebral Cortex},
  volume={28},
  number={9},
  pages={3095--3114},
  year={2018},
  publisher={Oxford University Press}
}

@article{KDE,
author = {Weglarczyk, Stanislaw},
year = {2018},
month = {11},
pages = {00037},
title = {Kernel density estimation and its application},
volume = {23},
journal = {ITM Web of Conferences}
}

@article{MARRON1988195,
title = {Canonical kernels for density estimation},
journal = {Statistics \& Probability Letters},
volume = {7},
number = {3},
pages = {195-199},
year = {1988},
author = {J.S. Marron and D. Nolan}
}

@ARTICLE{DTW,
  author={Sakoe, H. and Chiba, S.},
  journal={IEEE Transactions on Acoustics, Speech, and Signal Processing}, 
  title={Dynamic programming algorithm optimization for spoken word recognition}, 
  year={1978},
  volume={26},
  number={1},
  pages={43-49}
}

@article{wasserstein_ref,
  author  = {Yossi Rubner and Carlo Tomasi and Leonidas J. Guibas},
  title   = {The Earth Mover's Distance as a Metric for Image Retrieval},
  journal = {International Journal of Computer Vision},
  volume  = {40},
  number  = {2},
  pages   = {99--121},
  year    = {2000}
}

@article{szekely_energy,
  author  = {G{\'a}bor J. Sz{\'e}kely and Maria L. Rizzo},
  title   = {Energy Statistics: A Class of Statistics Based on Distances},
  journal = {Journal of Statistical Planning and Inference},
  volume  = {143},
  number  = {8},
  pages   = {1249--1272},
  year    = {2013}
}

@article{gibbs_su,
  author  = {Gibbs, Alison L. and Su, Francis Edward},
  title   = {On Choosing and Bounding Probability Metrics},
  journal = {International Statistical Review},
  volume  = {70},
  number  = {3},
  pages   = {419--435},
  year    = {2002}
}

@article{taubetabiom,
  title     = {Amyloid-$\beta$ and Phosphorylated Tau are the Key Biomarkers and Predictors of Alzheimer’s Disease},
  author    = {Jangampalli Adi, Pradeepkiran and Baig, Javaria and Islam, Md Ariful and others},
  journal   = {Aging and Disease},
volume = {16},
  number = {2},
  pages = {658-682},
  year      = {2025}
}

@article{univariate1,
  author    = {Park, Sang Won and Yeo, Na Young and Lee, Jinsu and others},
  title     = {Machine learning application for classification of Alzheimer's disease stages using $^{18}$F-flortaucipir positron emission tomography},
  journal   = {BioMedical Engineering OnLine},
  volume    = {22},
  pages     = {40},
  year      = {2023}
}

@article{univariate2,
title = {Classification of Alzheimer’s disease using MRI data based on Deep Learning Techniques},
journal = {Journal of King Saud University - Computer and Information Sciences},
volume = {36},
number = {2},
pages = {101940},
year = {2024},
author = {Shaymaa E. Sorour and Amr A. Abd El-Mageed and Khalied M. Albarrak and others},
}

@article{CDRSB,
  title   = {Sum of boxes of the clinical dementia rating scale highly predicts conversion or reversion in predementia stages},
  author  = {Tzeng, Ray-Chang and Yang, Yu-Wan and Hsu, Kai-Cheng and others},
  journal = {Frontiers in Aging Neuroscience},
  volume  = {14},
  pages   = {1021792},
  year    = {2022}
}

@article{ADAS-cog,
  title   = {The Alzheimer's Disease Assessment Scale-Cognitive Subscale (ADAS-Cog): Modifications and Responsiveness in Pre-Dementia Populations. A Narrative Review},
  author  = {Kueper, Jacqueline K. and Speechley, Mark and Montero-Odasso, Manuel},
  journal = {Journal of Alzheimer's Disease : JAD},
  volume  = {63},
  number  = {2},
  pages   = {423--444},
  year    = {2018}
}

@article{MMSE,
  author       = {Arevalo-Rodriguez, Ingrid and Smailagic, Nadja and Roqu{\'e} I Figuls, Marta and Ciapponi, Agust{\'i}n and Sanchez-Perez, Erick and Giannakou, Antri and Pedraza, Olga L. and Bonfill Cosp, Xavier and Cullum, Sarah},
  title        = {Mini-Mental State Examination ({MMSE}) for the Detection of Alzheimer's Disease and Other Dementias in People with Mild Cognitive Impairment ({MCI})},
  journal      = {Cochrane Database of Systematic Reviews},
  volume       = {2015},
  number       = {3},
  pages        = {CD010783},
  year         = {2015},
  publisher    = {John Wiley \& Sons, Ltd}
}

@article{RAVLT,
author = {Carone, Dominic},
year = {2007},
pages = {62-63},
title = {E. Strauss, E. M. S. Sherman, \& O. Spreen, A Compendium of Neuropsychological Tests: Administration, Norms, and Commentary},
volume = {14},
journal = {Applied Neuropsychology}
}

@book{BrainLobes,
  title     = {Cognitive Neuroscience: The Biology of the Mind},
  author    = {Gazzaniga, Michael S. and Ivry, Richard B. and Mangun, George R.},
  edition   = {5th},
  year      = {2018},
  publisher = {W. W. Norton \& Company}
}

@article{Yeo2011,
  title     = {The organization of the human cerebral cortex estimated by intrinsic functional connectivity},
  author    = {Yeo, B. T. Thomas and Krienen, Fenna M. and Sepulcre, Jorge and others},
  journal   = {Journal of Neurophysiology},
  volume    = {106},
  number    = {3},
  pages     = {1125--1165},
  year      = {2011}
}

@INPROCEEDINGS{multimodal,
  author={Thanh, Vu Duy and Trung Le, Thanh and Tuan, Pham Minh and Linh Trung, Nguyen and Abed-Meraim, Karim and Adel, Mouloud and Dung, Nguyen Viet and Thanh Trung, Nguyen and Long, Dinh Doan and Chén, Oliver Y.},
  booktitle={2024 International Conference on Multimedia Analysis and Pattern Recognition (MAPR)}, 
  title={Tensor Kernel Learning for Classification of Alzheimer’s Conditions using Multimodal Data}, 
  year={2024},
  volume={},
  number={},
  pages={1-6}
  }

@article{Yuan2012Multi,
  author    = {Lei Yuan and Yalin Wang and Paul M. Thompson and Vaibhav A. Narayan and others},
  title     = {Multi-source feature learning for joint analysis of incomplete multiple heterogeneous neuroimaging data},
  journal   = {NeuroImage},
  year      = {2012},
  volume    = {61},
  number    = {3},
  pages     = {622--632}
}

@Article{univariate4,
  author       = {Bueno-Cayo, Alma M. and del Rio Carmona, Minerva and Castell-Enguix, Rosa and others},
  title        = {Predicting Scores on the Mini-Mental State Examination (MMSE) from Spontaneous Speech},
  journal      = {Behavioral Sciences},
  volume       = {12},
  number       = {9},
  pages        = {339},
  year         = {2022}
}

@article{univariate6,
  author    = {Lin, Qi and Rosenberg, Monica D. and Yoo, Kwangsun and others},
  title     = {Resting-State Functional Connectivity Predicts Cognitive Impairment Related to Alzheimer's Disease},
  journal   = {Frontiers in Aging Neuroscience},
  volume    = {10},
  pages     = {94},
  year      = {2018}
}

@article{SanabriaDiaz2013,
  author    = {Sanabria-Diaz, Gretel and Martínez-Montes, Eduardo and Melie-Garcia, Lester},
  title     = {Glucose metabolism during resting state reveals abnormal brain networks organization in the Alzheimer's disease and mild cognitive impairment},
  journal   = {PLoS ONE},
  volume    = {8},
  number    = {7},
  pages     = {e68860},
  year      = {2013},
  publisher = {Public Library of Science}
}

@article{Mosconi2006,
  author    = {Lisa Mosconi and Sandro Sorbi and Mony J. de Leon and others},
  title     = {Hypometabolism exceeds atrophy in presymptomatic early-onset familial Alzheimer's disease},
  journal   = {Journal of Nuclear Medicine},
  year      = {2006},
  volume    = {47},
  number    = {11},
  pages     = {1778--1786}
}
%





\end{document}